___________________________________________________________________

# Multi-focus thermal image fusion

**Radek Benes[*1], Pavel Dvorak[*1], Marcos Faundez-Zanuy[*2], Virginia Espinosa-Duró[*2], Jiri Mekyska[*1]**

[*1] Department of Telecommunications, Faculty of Electrical Engineering and Communication, Brno University of Technology, Purkynova 118, 612 00 Brno, Czech Republic

[*2] EUP Mataró, Tecnocampus, 08302 Mataro, Barcelona, Spain

**ABSTRACT**

This paper proposes a novel algorithm for multi-focus thermal image fusion. The algorithm is based on local activity analysis and advanced pre-selection of images into fusion process. The algorithm improves the object temperature measurement error up to 5ºC. The proposed algorithm is evaluated by half total error rate, root mean squared error, cross correlation and visual inspection. To the best of our knowledge, this is the first work devoted to multi-focus thermal image fusion. For testing of proposed algorithm we acquire six thermal image set with objects at different focal depth.

## 1. Introduction

In an image only those objects within the depth of field of the camera are focused, while other objects are blurred. To obtain an image with every object in focus, we usually need to fuse the images taken from the same view point under different focal settings.

The aim of image fusion is to integrate complementary and redundant information from multiple images to create a composite that contains a "better" description of the scene than any of the individual source images [Huang et Jing 2007], [Eltoukhy2003]. The image fusion can be also assumed as an approach for de-noising of images [Cosmin2012]

Image fusion plays important roles in many different fields such as remote sensing, biomedical imaging, computer vision and defense system. While this topic has been carefully studied in the visible range (300-700 nm), to be best of our knowledge there is no work with thermal spectrum range (3-14 μm).

Nowadays, there are several different approaches for multi-focus, image fusion visible range. Mainly, they are divided into two groups according to the domain in which they work. [Shutao 2011] One group works in the spatial domain and computes focus in particular parts of the image directly from the source data. [Shutao 2008] The other group works in transform domain, usually wavelet, curvelet, contourlet, fourier or pyramid transform. [Shutao 2011][Denipote 2008] In the first case, focus rate is often measured using first or second derivative. In case of first derivative, gradient value is computed in each pixel second derivative is computed by Laplacian, or their modifications. In [Huang et Jing 2007], [Madhavi 2011] and [Shutao 2011] focus rate is also called activity level. When it is computed, images have to be combined. To compute a particular pixel of final image, only a pixel from the best focused image in this point can be copied. But a more common way is to weight  this point in every image according to the activity level. The resulting point is a weighted average [Blum

___________________________________________________________________________

2005]. There are also certain advanced methods that use neural networks [huang2007][li2002] or a multiscale approach [liu2001][zhang1999].

## 1.1 Measurement of quality of image fusion

In this paper, we present several experiments that point out the blurring effect on accuracy of temperature as measured in thermal images; and, we propose an algorithm for image fusion that alleviates the temperature errors due to blurring.

There are subjective and objective measures for fusion quality evaluation. Objective methods are more common because of their low computational burden [Shutao 2011]. These methods are divided into two groups. One group needs a reference image, which is usually not present, while the other group does not need one. In thermal imaging it is usually impossible to directly acquire a reference image due to the small focal length of thermal cameras. Therefore, for evaluation purposes we have to manually create a reference image.

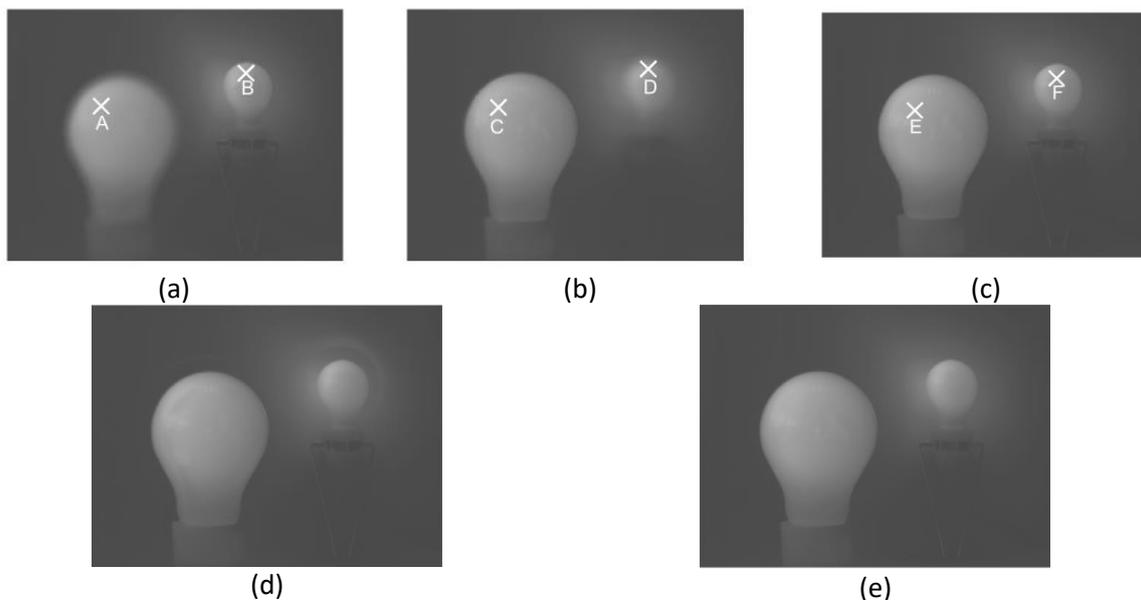

Figure 1: (a) One image from image set, where more distant bulb is focused. (b) one image from image set, where closer bulb is focused. (c) manually-fused image – both bulbs are focused. (d) image fused from all 96 images, (e) image fused from reduced number of images (glare around the bulb is lower).

In this paper, all image sets contain two objects in two different distances, so they are focused in different images (see Fig. 1 (a) and (b)). In testing datasets, both objects are situated in the scene so that they are not overlapping. This restriction was defined only to simplify the creation of a reference image. However, , for further utilization, the objects can be arbitrarily overlapped. The reference image is created manually from two input images (each one contains the best-focused first or second object). Because both objects are not overlapped the reference image (e.g. Fig. 1 (c)) is created by a

___

subjective selection of focused pixels from the second image (e.g. left bulb from Fig. 1 (b)) and their incorporation into first image (e.g. Fig. 1 (a)).

Methods for fusion quality evaluation were chosen from [Madhavi 2011] [Blum 2005] [Chen 2005] [Eskoliciou1995]. A few methods requiring reference image were used in this work and those are the following

1. The correlation (CC)

$$CC = \frac{2 \cdot \sum_{i=1}^{N} \sum_{j=1}^{M} R(i,j) F(i,j)}{\sum_{i=1}^{N} \sum_{j=1}^{M} R(i,j)^2 + \sum_{i=1}^{N} \sum_{j=1}^{M} F(i,j)^2}, \quad (1)$$

where *N*, *M* denote the size of image, *R* and *F* are the reference and fused image, respectively.

2. The Root Mean Square Error (RMSE)

$$\text{RMSE} = \sqrt{\frac{1}{NM} \sum_{i=1}^{N} \sum_{j=1}^{M} |R(i,j) - F(i,j)|^2}, \quad (2)$$

where N, M are the size of image.

3. The Mean Absolute Error (MAE)

$$MAE = \frac{1}{N} \sum_{i=1}^{N} |f_i - y_i|, \quad (3)$$

where $f_i$ is predicted and $y_i$ is computed value.

## 2. Materials and methods

### 2.1 Blurring effect on temperature measurement accuracy

Image blurring in visible images provides a degradation of the quality of the image. Edges do not appear sharp and the objects lose their details and their identification is more difficult. The absolute value of a single pixel in a visible image is related to the object itself as well as the illumination. When comparing two images with the same relative value between neighboring pixels the different absolute value is interpreted as an increase of uniform illumination. Within a reasonable range, this does not affect the ability to interpret the image content.

In thermal images the thermal camera acquires an absolute temperature value, which is related to the object itself and it is not related to the illumination of the scene, which is irrelevant. Therefore, the absolute value of the pixel is important as it correlates to measured temperature of the object. Thermal cameras provide accurate measurements when the object is focused. The problem appears when there are different objects at different focal distances. For the sake of simplicity we will consider that there are only two objects and both objects are at the same temperature. However, it is straightforward to generalize for higher numbers of objects and/or different temperatures.

Considering a scene that contains two objects at different focal distances and with the same temperature (T), we have to choose between focusing the image for the object 1 or the object 2. When focusing the camera for object one (more distant bulb in Figure (a)) the temperature $T_{object1}$ will be measured accurately but temperature $T_{object2}$ will be measured with error

$$T_{object1} = T, T_{object2} = T + error_1. \quad (4)$$

______________________________________________________________________________

When focusing the camera for object two (closer bulb in Figure (b)), the situation is similar – one temperature is measured with error and second is measured precisely.

$$T_{object1} = T + error_2, T_{object2} = T. \qquad (5)$$

The ideal situation would be to obtain the correct temperature on both objects simultaneously but this is not possible without using a multi-focus image fusion. In this case, we expect that the total error in the fused image will be smaller than for each of the input images alone, being each of these images the acquisition at one focal distance.

Thus, for this two objects problem we will define the half total error as:

$$HTE = (error_1 + error_2)/2. \qquad (6)$$

A desirable property of a multi-focus fused image is $HTE \cong 0$, and of course, smaller value than *HTE* of each of the fused images.

## 2.2 Proposed algorithm

Multi-focus image fusion is a process that combines multiple images captured with different depth of field into a single image. A Block diagram of multi-focus fusion is depicted in Figure . The Multi-focus image fusion process starts with analysis of input images and ends with the combination of these images into a single image. This combined image is more focused than each of the input images alone.

Often the number of images that enter the fusion process is very high and some of them do not contain any useful information because there are completely blurred. Therefore we propose the method for reducing the number of images. This image pre-selection process also reduces the computational burden of the algorithm.

We try to compare two approaches. One is simple image fusion performed directly with all input images; the second one contains automatic reduction of images entering into the image combination (Figure ). Since totally blurred images generate noise into the fused image, the performance of the fusion process decreases if fusion is performed directly with all images.

This whole fusion system is described in next sections, and consists of the following steps:

1. Measurement of the activity level
2. selection of the best images for fusion,
3. combination of the selected images.

_________________________________________________________________________________

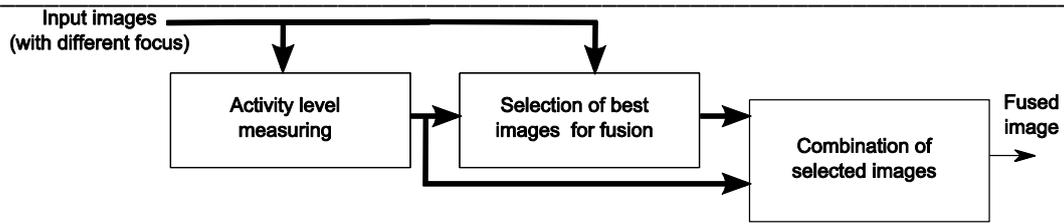

Figure 2: Block diagram of the proposed fusion (with reduction of input images)

In our image sets, there are 96 images in each image set and each image is captured with different focus positions of a thermal camera (some of them are depicted in Figure (a) – (d)). Some images contain sharp objects, but many of them are completely blurred. This shows the importance of the reduction of input images.

### Activity level (AL) measuring

In our previous paper [Faundez 2011] we compared some approaches for image focus measuring but they were used for whole image. In this work we will use a local measure, because we want to combine different parts of different images on a local basis. According to results presented in [Faundez 2011], the energy of Laplacian has been selected to compare information content in particular parts of images.

First of all, the information content in local parts of images must be computed. In [Huang et Jing 2007], [Madhavi 2011], [Shutao 2011], this information content is called "Activity Level" (AL). This AL is a measure of information in every pixel of each image. The activity level can be computed in different ways.

The energy of Laplacian in point $(x, y)$ can be computed according to formula

$$\text{EOL}(x, y) = \left(f_{xx} + f_{yy}\right)^2, \quad (7)$$

where:

$$\begin{aligned}(f_{xx} + f_{yy}) = &-I(x-1, y-1) - 4I(x-1, y) - I(x-1, y+1) - 4I(x, y-1) + 20I(x, y) \\ &- 4I(x, y+1) - I(x+1, y-1) - 4I(x+1, y) - I(x+1, y+1).\end{aligned}$$

The activity level measure $m(x, y)$ in point $(x, y)$ can be computed from $\text{EOL}(x, y)$ [Subbarao and Tyan, 1998]. It is computed as average value of EOL in certain neighborhood multiplied by variance in the same neighborhood.

$$m(x, y) = \overline{\text{EOL}(x, y)} \cdot \sigma^{(x,y)^2}, \quad (8)$$

where $\overline{\text{EOL}(x, y)}$ is the averaged value of EOL in neighborhood of pixel $(x, y)$ of size $w$

______________________________________________________________________

$$\overline{\mathrm{EOL}(x,y)} = \frac{1}{w \cdot w} \sum_{i=x-w/2}^{x+w/2} \sum_{j=y-w/2}^{y+w/2} \left(\mathrm{EOL}(i,j)\right), \tag{9}$$

and $\sigma(x,y)^2$ is variance in the same neighborhood

$$\sigma(x,y)^2 = \frac{\sum_{i=x-w/2}^{x+w/2} \sum_{j=y-w/2}^{y+w/2} (\mathrm{EOL}(i,j) - \overline{\mathrm{EOL}(x,y)})}{w \cdot w}. \tag{10}$$

With this equation, activity level can be computed in all pixels of all images. In further computation, these values of activity level are weights that are used for combining input images. Examples of activity level matrices are depicted in Figure . Before the combination of images, it is suitable to select a subset of images entering to fusion process.

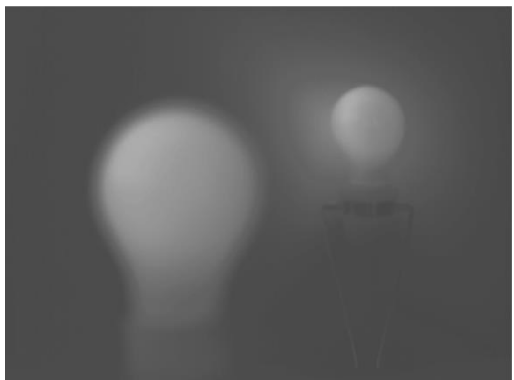

(a)

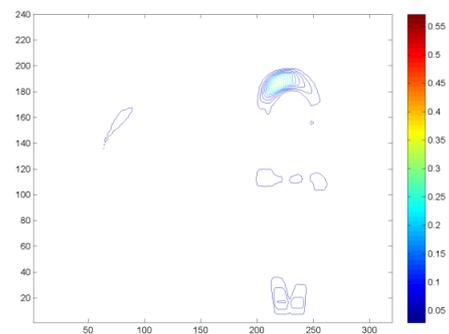

(e)

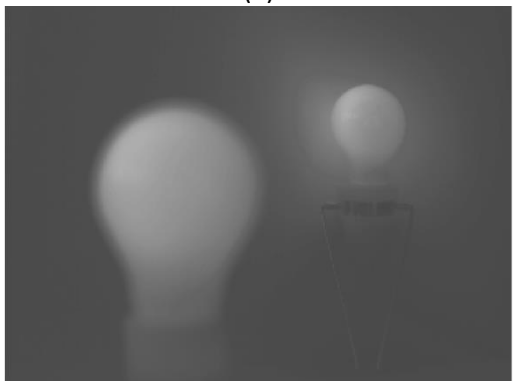

(b)

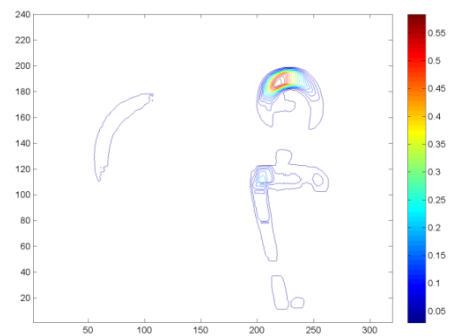

(f)

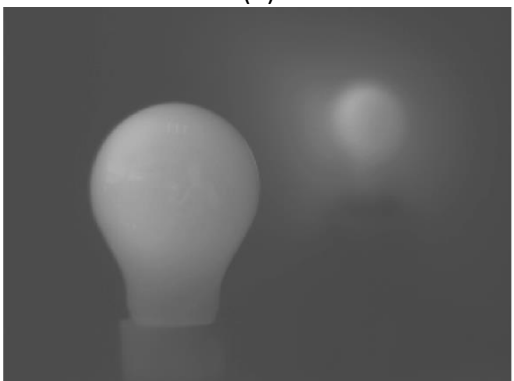

(c)

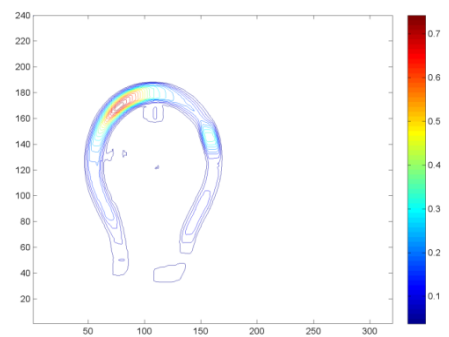

(g)

___

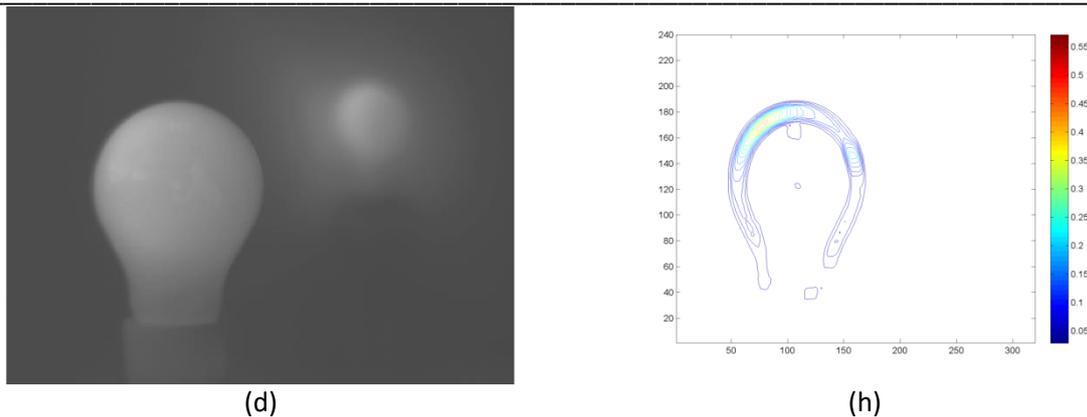

(d)                                           (h)

Figure 3: Figures (a) – (d) depicts selected thermal images form the bulb dataset. Corresponding activity levels are depicted in figures (e) – (h).

## Selection of the best images for fusion

AL is computed for each input image. Then the maximum of activity level in each image is found according to

$$m_{i,\max} = \max_{x,y}(m_i(x,y)), \qquad (11)$$

where $m_i(x,y)$ is AL in $i$-th image.

In blurred images, the activity level and also its maximum are very small. The dependency of maximal value of activity level on the image number can be plotted into a graph. Figure (a) shows this situation for image set 3, which contains two bulbs at different focal distances.

The selection of suitable images for fusion is straightforward. The algorithm finds all peaks and takes into account only images that are around these peaks. The appropriate number of selected images around each peak was found experimentally. The algorithm was tested with different setup and according to RMSE the most suitable number of images around each peak was selected. The dependence of RMSE on number of images taken into account around each peak is depicted in Figure (b). One can see that the most suitable number of images selected around each peak is 4.

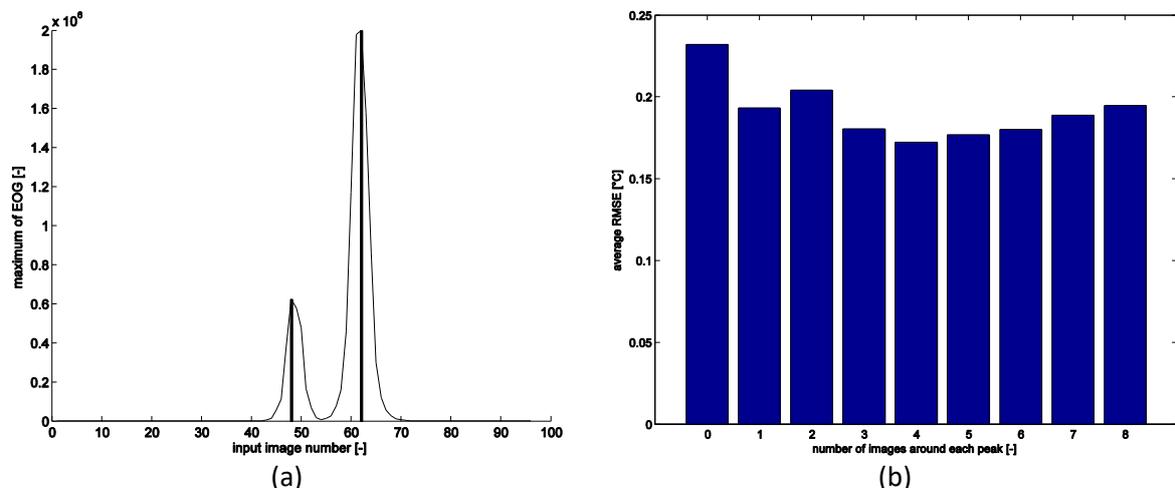

(a)                                           (b)

Figure 4: (a) Dependency of maximal value of activity level versus image number for image set 3. The peaks are highlighted with a straight bar. (b) Dependency of RMSE on the number of images selected around each peak.

______________________________________________________________________________

### Combination of input images

For the combination of input images, a simple method based on "pixel-level weighted averaging" is employed. The activity level is used as the weighting for combining images. For the correct combination of images, the sum of weights in each pixel has to be 1. Thus the weights have to be normalized in the following way:

$$m_i'(x,y) = \frac{m_i(x,y)}{\sum_{i=1}^{N} m_i(x,y)}. \qquad (12)$$

Then, the linear combination of input images can be expressed with the equation for the fused image:

$$I_f(x,y) = \sum_{i=1}^{N} m_i'(x,y) \cdot I_i(x,y), \qquad (13)$$

where $I_f(x,y)$ is the fused image and $I_i(x,y)$ are the input images.

## 3. Results

In this section we present the experimental results obtained with the database described in section 2. We compare the method with proposed block "reduce number of images" (see Figure ) and the method without this block (simple case of linear fusion). For evaluation the measurement of HTE, RMSE, cross correlation and visual comparison have been used. Some of these evaluation methods require reference image, thus "manually fused image" has been constructed. This was not difficult because in testing image sets here, there are only two objects in two different depths.

### 3.1 Database

In order to develop a thermal image fusion system we require a database of thermal images. These images should contain several objects at different distances. However, thermal cameras do not provide the several million pixels resolution provided by visible ones. Thus, we will deal with the resolution 320x240 px.

We used a thermographic camera TESTO 882-3 equipped with an uncooled detector and a spectral sensitivity range from 8 to 14 μm. It has a removable German optic lens with these main features:

- image resolution: 320 x 240 px,
- spectral sensitivity: 8 to 14μm,
- thermal sensitivity (NETD) <0.06 ºC at 30 ºC,
- geometric resolution (IFOV): 1,7 mrad,
- detector type: silicon microbolometer uncooled, temperature stabilized.
- FOV: 32°x23°; focal distance: 15mm; fixed aperture: f/0,95

The database consists of six image sets. In each set, the camera acquires one image of the scene at each lens position. In our case we have manually moved the lens in 1 mm steps, which provides a total of 96 positions. Thus, each set consists of 96 different images of the one scene. For this purpose, we have attached a millimeter tape to the objective. We also used a stable tripod in order

___

to acquire the same scene for each scene position and a dimmer to fix the bulb current. This is the same as in previous paper [Faundez 2011] although there we used a TESTO 880-3 thermal camera that provides lower resolution (160x120) and sensitivity (NETD < 0.1 ºC).

We have acquired six image sets:

1. Image set 1 (Figure 5 (a)): scene is made up of mobile phone and RS-232 interface in different distances and homogenous heat absorbing background. Distance between camera and the first object is 35 cm and its temperature is 41.2 °C. The distance between objects is 40 cm for all images sets. The maximum temperature of the second object is 32.9 °C.
2. Images set 2 (Figure (b)): scene is made up of mobile phone and RS-232 interface in different distances and homogenous heat absorbing background. Distance between camera and the first object is only 15 cm and its temperature is 39.4 °C. The maximum temperature of the second object is 55.9 °C.
3. Image set 3 (Figure (c), selected images are depicted in Figure (a) – (d)): scene is made up of two bulbs in different distances and non-homogenous background (partially black and partially white). The bulbs are acquired with a view to the holders. Distance between camera and the first object is 30 cm as in all bulb image sets. The temperature of $1^{st}$ bulb is 51.7 °C and $2^{nd}$ is 50.4 °C.
4. Image set 4 (Figure (d)): scene is made up of two bulbs in different distances and homogenous white background. The bulbs are acquired with a view to the holders. The temperature of the first bulb is 43.3 °C and $2^{nd}$ is 41.3 °C.
5. Image set 5 (Figure (e)): scene is made up of two bulbs in different distances and homogenous white background. The bulbs are acquired without a view to the holders. The temperature of the first bulb is 57.0 °C and $2^{nd}$ is 53.6 °C.
6. Image set 6 (Figure (f)): scene is made up of two bulbs in different distances and homogenous heat absorbing black background. The bulbs are acquired with a view to the holders. The temperature of the first bulb is 57.9 °C and $2^{nd}$ is 54.7 °C.

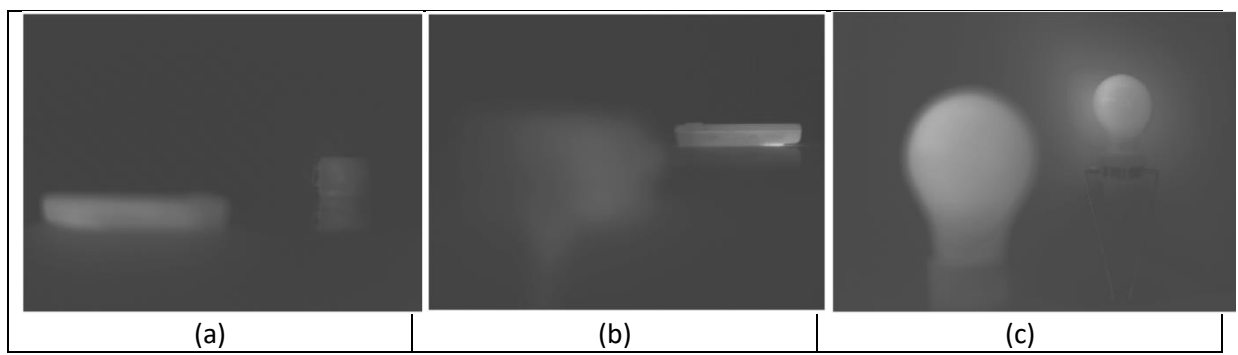

|    (a)    |    (b)    |    (c)    |

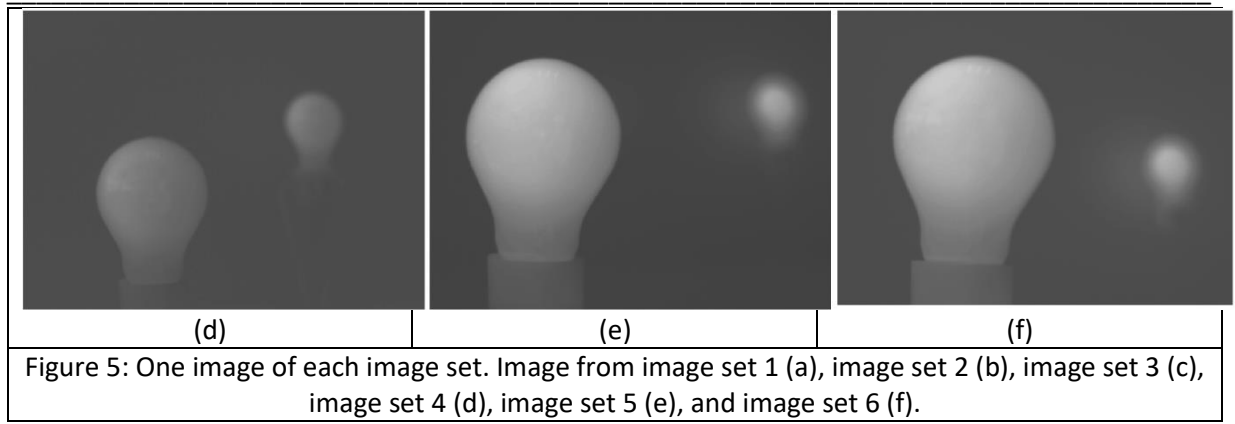

Figure 5: One image of each image set. Image from image set 1 (a), image set 2 (b), image set 3 (c), image set 4 (d), image set 5 (e), and image set 6 (f).

### 3.2 Measure of HTE

This measurement is based on the fact, that in a case when thermal camera is not focused on the object, the temperature of the object is measured with error. For this evaluation, two images (each focused on one object (see Figure (a), (b))) has been selected. In Figure (a), where the camera is focused on the more distant object, the temperature of this far object is measured precisely (measured in point B), but the temperature of the closer object (measured in point A) is measured with error $e_A$. At this point the average error $HTE_1 = \frac{e_A+0}{2}$ can be measured. Inverse situation is shown in Figure (b), where the average error can be computed as $HTE_2 = \frac{e_D+0}{2}$, because of the temperature of the far object (point D) is measured with error $e_D$.

Similar error can be measured in the fused image (Figure (c)). In this situation, both temperatures in the fused image are slightly inaccurate (points E, F in Figure (c)), but thanks to fusion, these errors are not so significant. Let $e_F$ be the error of the far object and $e_E$ error of the near object. Since two methods are compared in this article – without and with the reduction of number of input images, two result images exist and the average error for both of them is computed as $HTE_3 = \frac{e_E+e_F}{2}$ and $HTE_4 = \frac{e_{E'}+e_{F'}}{2}$, respectively.

These errors, $HTE_1, HTE_2, HTE_3, HTE_4$, can be measured automatically for each image set. The absolute values of these errors are summarized in Figure 1 (a). The statistic parameters (the sample minimum, lower quartile, median, upper quartile and sample maximum) of these errors can be seen in so-called box plot in Figure 1 (b). It can be observed that errors $HTE_1, HTE_2$ are significant. This is due to the inaccuracy that comes from bad focus of the camera. Thanks to fusion, the HTE error is lower and moreover if the image number reduction is done, the HTE is even lower (represented in Figure 1 (a) and Figure 1 (b) as $HTE_4$).

It is important to observe that the inaccuracies in temperature measurement for blurred objects can be as high as 5ºC. While this error can be neglected for some industrial applications, it can be considerable for biometric recognition of people, thermal isolation analysis, etc.

____________________________________________________________________________________

### 3.3  Visual comparison (bulb image-set)

If the image is fused without a proposed block for reduction of images, one can see the glare (Figure (d) around bulbs which is generated by the redundant images. Conversely, when the proposed block is used, the glare is reduced (Figure 1 (e)) and the fusion is clearer and more precise.

### 3.4  RMSE computed for a whole image

We have also evaluated the proposed algorithm using the Root Mean Squared Error (RMSE). The RMSE is computed between the manually fused image and the image fused with algorithm. The results for different image sets and for both cases (without image reduction vs. with image reduction) are summarized in Tab. 1. These values are statistically processed and depicted in a form of boxplot in Figure 1 (c). We can observe that the proposed method with image number reduction provides smaller errors than the combination of all the images.

Tab. 1: RMSE measured for all available image sets.

| Image set number | RMSE computed between fused image (full number of images used) and manually fused image | RMSE computed between fused image (reduced number of images used) and manually fused image |
| --- | --- | --- |
| 1 | 0.2202 | 0.1803 |
| 2 | 0.2764 | 0.2583 |
| 3 | 0.3014 | 0.1999 |
| 4 | 0.1488 | 0.1342 |
| 5 | 0.4491 | 0.2648 |
| 6 | 0.4438 | 0.3307 |

### 3.5  Cross correlation (CC) for a whole image

Similarly to RMSE, the cross correlation (CC) is computed for the whole image. Figure 1 (d) shows the statistical values for all the datasets.

### 3.6  Comparison of fused image against all images entering into the fusion process

Values of RMSE and CC between fused image and each image entering into the fusion can be measured. These values are plotted in Figure 1 (e) and Figure 1 (f). One can see that the RMSE is the smallest in the case where one of the objects is focused. For the comparison, there are plotted lines that show RMSE computed for result images (without vs. with image selection) in Figure 1 (e). Similarly Figure 1 (f) shows CC. One can see that CC in fused image is significantly higher than CC computed between each input image and fused image.

_________________________________________________________________

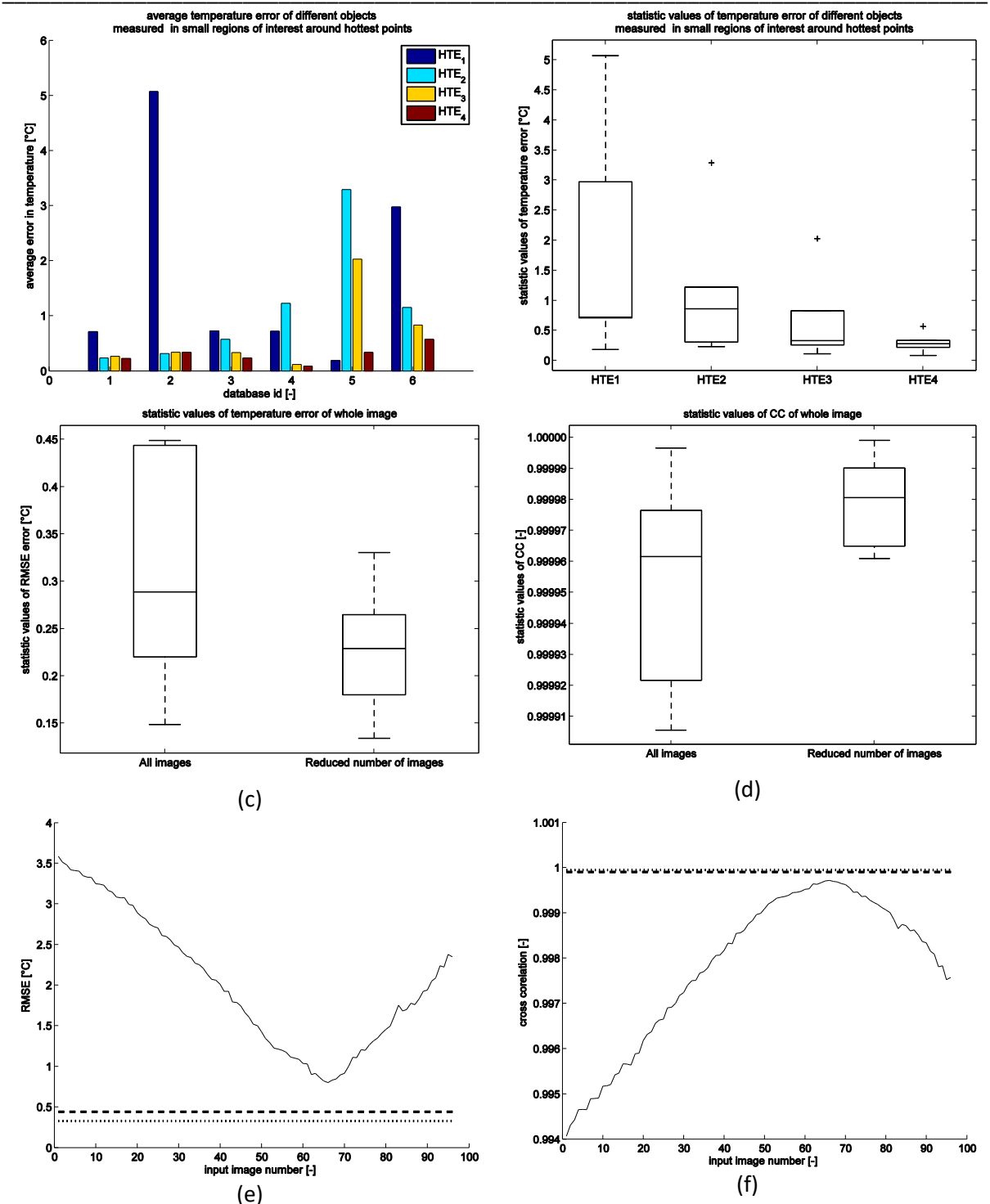

Figure 1: (a) Absolute values of HTE in particular image set. (b) Statistic values of HTE errors computed in whole databases. Statistical values of RMSE (c) and CC (d) for different fusion methods (performed with all images or with reduced number of images). Solid curve – RMSE (e) or CC (f) computed between *i*-th image in image set and manually fused image. Dashed line (for comparison) – RMSE (e) or CC (f) measured between image fused without image number reduction and manually fused image. Dotted line (for comparison) – RMSE (e) or CC (f) measured between image fused with proposed method and manually fused image.

___________________________________________________________________________

**Conclusions**

In this paper we have proposed a method for an image fusion in the thermal spectrum. Moreover, we have proposed its improvement through a pre-selection of images entering into the fusion process. The quality of the proposed pre-selection method has been verified by measurements of these quality and error parameters: half total error (HTE), root mean squared error (RMSE), and cross correlation (CC). The results of fusion with proposed pre-selection are compared with a standard fusion method. Achieved results are described in detail in section 3. The smaller values of error parameters (HTE and RMSE) represent the higher quality of fusion. The pre-selection decreases the average HTE value (averaged in acquired image sets) from 0.326 [°C] to 0.271 [°C]. Also the average RMSE value decreases from 0.31 to 0.22. For example in the fifth dataset, the reduction of RMSE is about 42%. Vice-versa for the cross correlation parameter where the higher value indicates the higher quality of fusion. Thanks to proposed image pre-selection the CC increases from 0.999962 to 0.999980. All performed measurements prove that proposed fusion method and its improvement significantly decreases the temperature errors.

For testing purposes the database of thermal images has been acquired using thermal camera TESTO 882-3. The database consists of six image sets of complex objects with different temperatures and in different distances from the thermal camera. Each image set consist of 96 images manually acquired with different focus.

The results prove that a multi-focus image fusion can alleviate the errors in temperature measurements. This has been checked using the acquired datasets. The quantitative parameters, HTE , RMSE, and CC were measured and, in addition, human inspection was performed.

**Acknowledgements**

This work has been supported by FEDER and MEC, TEC2009-14123-C04-04. We also want to acknowledge project KONTAKT-ME 10123, project SIX (CZ.1.05/2.1.00/03.0072), project VG20102014033, project CZ.1.07/2.3.00/20.0094 and project GACR 102/12/1104.

### 3.1 APPENDIX

Just as in visible image acquisition systems, an optical system capable of focusing all rays of light from a point in the object plane to the same point in the focal plane is desired. The same goal is also desired when dealing with thermal imagers in order to get clear and focused images in thermal infrared spectrum. However, all kind of lens aberrations as well as deviations due to diffraction, drastically reduce the capability to focus the image of interest.

Chromatic aberration, as the most critical aberration aspect when dealing with broadband spectrum images (which is the case of MWIR and LWIR thermal images) will be described below followed by the description of a diffraction effect. The problem becomes more difficult to solve when more than one object ,located at different distances in the scene, requires to be brought into focus due to the constraints in depth of field design.

_____________________________________________________________________________

## Chromatic aberration.

Chromatic aberration [Jac00] is an undesirable optical effect that promotes the inability of the lens to focus all the colors (different wavelengths) at the same focal point. This effect is due to the spread dispersion phenomenon concerning the refractive index variation with wavelength. Normal lens show normal dispersion, that is, the index of refraction $n$, decreases with increasing wavelength. Thus, the light beam with longer wavelength is refracted less than the shorter wavelength one. This behavior produces a set of different focal points.

In any case, the correction tasks in the visible spectrum are reasonable to achieve due to the short range of wavelengths to deal with. This is not the case in the MWIR and LWIR operating ranges, where thermal IR sensors measure simultaneously over broadband wavelength. Thus, while the change is 400nm between the violet and red end of the EM spectrum, in both the MWIR and LWIR spectra the wavelength ranges are 2000nm and 6000nm, respectively. Another challenging problem is the coupling between large wavelength and low refractive index. To promote the required refraction to deviate and to accuratelyconverge any infrared light comprised in the range, an IR transparent material with high refraction index is required for the design of lenses that might not otherwise be possible [Gre07].

## Diffraction Effect

Diffraction is an optical effect, which can limit the total resolution of any image acquisition process. Usually, light propagates in straight lines through air. However, this behavior is valid only when the wavelength of the light is much smaller than the size of the structure through which it passes. For smaller structures, such a gap or a small hole, which is the case of camera's aperture, light beams will suffer from a diffraction effect caused by a slight bending of light when it passes through such singular structures [conrad2].

Due to this effect, any image formed by a perfect optical lens of a point of light, do not correspond to a point, but to a circle called *Airy disc,* and determines maximum blur allowable by the optical system. Furthermore, the diameter of this circle will be used to define the theoretical maximum *spatial resolution* of the sensor and will be given by the following expression [Noah 2004]:

$$d = 2{,}44\lambda \frac{v}{D} \qquad (14)$$

where $\lambda$ is the light wavelength, $v$ is the distance from the image to the lens and $D$ is the effective aperture diameter. This equation can be generalized to Eq. 2 when the system is working slightly far off the minimum focal distance, and $N$ being the lens f number[Noah 2004]:

$$d = 2{,}44\lambda N \qquad (15)$$

Radek Benes, Pavel Dvorak, Marcos Faundez-Zanuy, Virginia Espinosa-Duró, Jiri Mekyska, Multi-focus thermal image fusion, Pattern Recognition Letters, Volume 34, Issue 5, 2013, Pages 536-544, ISSN 0167-8655, https://doi.org/10.1016/j.patrec.2012.11.011___

Typical available pixel sizes for MWIR and LWIR range from 20 to 50$\mu^2$, while less than 2$\mu^2$ may be found for visible spectrum.

This restriction seriously determines a closely related parameter called *depth of field* (DOF) - the range of distance that appears acceptably sharp in the resulting image. The DOF depends on three main parameters: aperture (f number), focus distance and focal length. [conrad1][conrad2]

Assuming a diffraction limited system, the DOF can be expressed as follows:

$$DOF = \frac{D^2}{4\lambda} \qquad (16)$$

The DOF is function only of the aperture diameter and the wavelength. According to the equation above, the *DOF decreases when wavelength increases.* This is the main difference between VIS and IR acquisition systems and uncovers why thermal imagers may not focus all planes of the acquired scene.

This assertion leads us to carry out the following proposed approach in an attempt to provide a novel solution to the described problem.

**Acknowledgement**

This work has been supported by FEDER and MEC, TEC2009-14123-C04-04.**References**

[Faundez 2011] Marcos Faundez-Zanuy, Jiří Mekyska, Virginia Espinosa-Duro,"On the focusing of thermal images" Pattern Recognition Letters, Volume 32, Issue 11, 2011, Pages 1548-1557, ISSN 0167-8655, https://doi.org/10.1016/j.patrec.2011.04.022

[Huang et Jing 2007] Wei Huang, Zhongliang Jing "Evaluation of focus measures in multi-focus image fusion" Pattern Recognition letters. Elsevier. Vol. 28 (2007) pp. 493–500.

[Madhavi 2011]R. Madhavi, K. Ashok Babu, "An all Approach for Multi-Focus Image Fusion Using Neural Network" International Journal of Computer Science and Telecommunications. Vol 2. Issue 8. (November 2011) pp. 23-29

[Blum 2005] R. S. Blum, Z. Xue, and Z. Zhang, "An Overview of Image Fusion", chapter in ''Multi-Sensor Image Fusion and Its Applications'', Editors: R. S. Blum and Zheng Liu. Taylor & Francis, 2005; 1-36. ISBN: 08493341792005

[Shutao 2011] Shutao Li, Bin Yang, and Jianwen Hu. 2011. "Performance comparison of different multi-resolution transforms for image fusion". *Inf. Fusion* 12, 2 (April 2011), 74-84.

[Chen 2005] Chen, Y.; Blum, R.S.; , "Experimental tests of image fusion for night vision," Information Fusion, 2005 8th International Conference on , vol.1, no., pp. 8 pp., 25-28 July 2005